\documentclass[conference]{IEEEtran}
\IEEEoverridecommandlockouts
\usepackage{cite}
\usepackage{amsmath,amssymb,amsfonts}
\usepackage{algorithmic}
\usepackage{graphicx}
\usepackage{textcomp}
\usepackage{xcolor}
\usepackage{cleveref}
\usepackage{url}
\usepackage{booktabs}
\def\BibTeX{{\rm B\kern-.05em{\sc i\kern-.025em b}\kern-.08em
    T\kern-.1667em\lower.7ex\hbox{E}\kern-.125emX}}
\begin{document}

\title{Deep Learning for Climate Action: Computer Vision Analysis of Visual Narratives on X\\
\thanks{This project was funded by BMBF project 16DKWN027b Climate Visions}
}

\author{
    \IEEEauthorblockN{Katharina Prasse, Marcel Kleinmann, Inken Adam, Kerstin Beckersjürgen, Andreas Edte, \\ Jona Frroku, Timotheus Gumpp, Steffen Jung}
    \IEEEauthorblockA{\textit{Data \& Web Science Group, University of Mannheim}\\
    Mannheim, Germany \\
    katharina.prasse@uni-mannheim.de}
    \\
    \IEEEauthorblockN{Isaac Bravo, Stefanie Walter}
    \IEEEauthorblockA{\textit{School of Governance, Technical University of Munich}\\
    Munich Germany}
    \\
    \IEEEauthorblockN{Margret Keuper}
    \IEEEauthorblockA{\textit{University of Mannheim and Max Planck Institute for Informatics, Saarland Informatics Campus}\\
    Mannheim and Saarbrücken, Germany}
}


\vspace{0.1cm}

\maketitle

\begin{abstract}
Climate change is one of the most pressing challenges of the 21st century, sparking widespread discourse across social media platforms. 
Activists, policymakers, and researchers seek to understand public sentiment and narratives while access to social media data has become increasingly restricted in the post-API era. 
In this study, we analyze a dataset of climate change-related tweets from X (formerly Twitter) shared in 2019, containing 730k tweets along with the shared images. 
Our approach integrates statistical analysis, image classification, object detection, and sentiment analysis to explore visual narratives in climate discourse. 
Additionally, we introduce a graphical user interface (GUI) to facilitate interactive data exploration. Our findings reveal key themes in climate communication, highlight sentiment divergence between images and text, and underscore the strengths and limitations of foundation models in analyzing social media imagery. 
By releasing our code and tools, we aim to support future research on the intersection of climate change, social media, and computer vision.
\end{abstract}

\begin{IEEEkeywords}
climate change, social media, computer vision, visual narratives
\end{IEEEkeywords}

\section{Introduction}
Climate change is a defining challenge of the 21st century.
Public discourse on this topic spans political debates, activism, and scientific communication, with social media playing a central role in shaping narratives. 
Platforms like X (formerly Twitter) provide a dynamic space where individuals, organizations, and policymakers share opinions, mobilize support, and react to events in real-time. 
Understanding these discussions is crucial for stakeholders aiming to engage with the public. 
While textual discourse analysis on social media is well established, the role of images remains understudied.
Visual content—ranging from protest photos and scientific infographics to memes and news reports—influences perception and engagement \cite{wang2018public, johann2023fridays, hayes2021greta}. 
Within social sciences, scholars have analyzed climate change imagery through thematic framing \cite{born2019bearing, Casas2019, hayes2021greta, mcgarry2024fire}, often relying on manual annotation \cite{mooseder2023social, o2023visual, schaefer2017frame}. 
Recent advancements have made large-scale visual analysis more accessible, enabling automated classification, sentiment analysis, and object detection.

In this work, we explore climate change discourse on X (formerly Twitter) through a computer vision lens. 
Using a dataset of 730k climate-related tweets from 2019, we analyze their images alongside their metadata to uncover key trends in image content, sentiment, and engagement. 

Our main contributions include:
\begin{enumerate}
    \item An image, text, and multi-modal analysis of climate change-related social media content, combining image classification, object detection, and sentiment analysis.
    \item The application of foundation models (e.g., Gemini, Moondream) alongside traditional deep learning architectures (e.g., ResNet, ViT, GroundingDINO) to assess their performance in visual climate discourse analysis.
    \item An interactive graphical user interface (GUI) for exploring tweet data and model predictions, supporting deeper qualitative analysis.
\end{enumerate}

By making our code and tools publicly available, we aim to facilitate further research at the intersection of climate change, social media, and computer vision.

\section{Related Work}

Analyzing climate change narratives on social media involves both textual and visual framing. 
In social sciences, framing theory is widely used to understand how climate issues are presented and interpreted \cite{goffman1974frame}.
Traditionally, frames have been studied in text-based content, but research on visual frames has gained traction recently e.g.~\cite{Casas2019, mooseder2023social, o2020more}.

\subsection{Visual Framing in Climate Change Communication}
Schäfer et al. \cite{schaefer2017frame} categorize visual frames into two types:
First, formal/stylistic frames, which focus on image composition and aesthetic properties.
Second, content-oriented frames, which analyze the subject matter and depicted image themes.
Most prior studies have focused on content-oriented frames \cite{Casas2019, mooseder2023social, o2020more,o2023visual, mcgarry2024fire, born2019bearing, hayes2021greta, rebich2015image, prasse2024spy}, using manual annotation to classify climate-related images. 
For example, Born et al. \cite{born2019bearing} explore how polar bears serve as an icon for climate change, while McGarry et al. \cite{mcgarry2024fire} analyze wildfire imagery. 
However, manual annotation is time-consuming and subjective, limiting the scalability of such studies. 
Recent work has sought to automate visual frame detection through unsupervised clustering \cite{prasse2024spy}, offering a more scalable alternative.

\subsection{Computer Vision for Social Media Analysis}
Advancements in computer vision and foundation models have enabled large-scale analysis of visual content. 
Vision-language models (VLMs) can now generate image descriptions, detect objects, and analyze sentiment with increasing accuracy \cite{gemini}. 
Prior research has leveraged deep learning to study climate-related images, using models such as:
ResNet and ViT for image classification of environmental topics \cite{resnet, vit}.
DETR and GroundingDINO for object detection in climate-related imagery \cite{detr, liu2025grounding}.
RoBERTa and LENS for sentiment analysis, bridging textual and visual understanding \cite{lens, tan2022roberta}.
Despite these advancements, sentiment analysis for images remains challenging, as models struggle with context-dependent sentiment (e.g., sarcasm, irony, or contrasting text-image relationships). 

\section{Data}
We accessed data from X (formerly Twitter) when the academic API granted free access to academia. 
General selection criteria have been chosen: The tweet must contain either "climate change", "climatechange", or "\#climatechange" and must be published in the year 2019.
In total, we have downloaded 730k tweets which contain at least one visual and the corresponding tweet data consisting of tweet text, date, author id, user reactions (e.g. likes, retweets, shares).
We disregarded tweets without images.

We created subsets of the image data for increased efficiency and sustainability of the analysis.
The first subset contains the 5k most liked tweets, whereas the second subset contains a random selection of 10k tweets to represent the whole data.
For sentiment analysis, we created a third labeled subset which contains 500 randomly sampled images from the first subset that are manually labeled for their text and image sentiment.
The annotations are given by two independent annotators, divergent labels are resolved during discussion.

\section{Methods}
The selection of methods is extensive: from task-specific models to general-purpose foundation models. 

\subsection{Frequency Analysis}
We provide a general overview using frequency analysis of the tweets. 
This includes word clouds of the tweet's hashtags to visualize the main themes present.
Moreover, we investigate overall emoji usage, tweet dates, and user engagement by visualization (e.g. histograms or line graphs).

\subsection{Foundation models}
We use two foundation models for image classification and sentiment analysis: 
(1) Gemini Pro \cite{gemini}, which architecture is based on the Transformer decoder, optimized for both text and image decoding. 
Post-training, fine-tuning, and reinforcement learning from human feedback further refine its capabilities.
(2) Moondream \cite{moondream} is a compact, open-source computer vision model developed by M87 Labs, designed specifically for answering questions about images. 
With only 1.6 billion parameters, Moondream is lightweight and optimized for deployment on a wide range of devices, including mobile phones and edge devices like Raspberry Pi. 

\subsection{Image Classification}
To detect the general topic of an image, we use image classification:
(1) ResNet \cite{resnet} models are employed, a common CNN architecture. 
More specifically, we use ResNet-18, which offers a good balance between performance and computational efficiency, while ResNet-50, provides greater depth and a higher capacity for capturing complex features.
(2) Vision transformers, i.e. ViT-Base-Patch16-224 \cite{vit}, leverage the transformer architecture originally designed for natural language processing tasks. 
ViTs split an image into patches and processes patches as a sequence, allowing it to capture long-range dependencies and global context.

\subsection{Object Detection}
We use two models for more intricate image understanding:
(1) DETR (DetrForObjectDetection) \cite{detr} utilizes a transformer architecture to detect and locate objects within an image. 
Image features are extracted using a ResNet-50 backbone pretrained on ImageNet, whereafter the model is trained on MS COCO \cite{mscoco}.
(2) GroundingDINO \cite{liu2025grounding} is an open-set object detector that combines the encoder-decoder model, DINO,  with large, refined pre-training. 

\subsection{Sentiment Analysis}
For the sentiment analysis task, we
selected three possible labels: "positive", "neutral", and
"negative" in line with the majority of works in this field:
(1) We employ the text-only model RoBERTa (Robustly Optimized BERT Pre-training Approach), which is an advancement over BERT \cite{liao2021improved}. 
While it maintains a similar architecture, RoBERTa employs dynamic masking during pre-training instead of static masking.
Additionally, it is only pre-trained on masked language modeling and not on next-sentence prediction.
(2) The vision model LENS (Large Language Models ENhanced to See) \cite{lens} is a modular approach designed to integrate vision capabilities into any off-the-shelf large language model (LLM) without requiring additional multi-modal training or data. 
It works by utilizing small, independent vision modules to generate detailed textual descriptions of images, which are then processed by the LLM to perform tasks like object recognition and VQA. 

\section{Analysis}
We report frequency analysis and sentiment analysis in the main paper and report object classficiation results in \Cref{sec:object}.
\subsection{Frequency Analysis}
\begin{figure}[ht]
    \centering
    \includegraphics[width=0.8\linewidth, trim={0 0 0 0.75cm },clip]{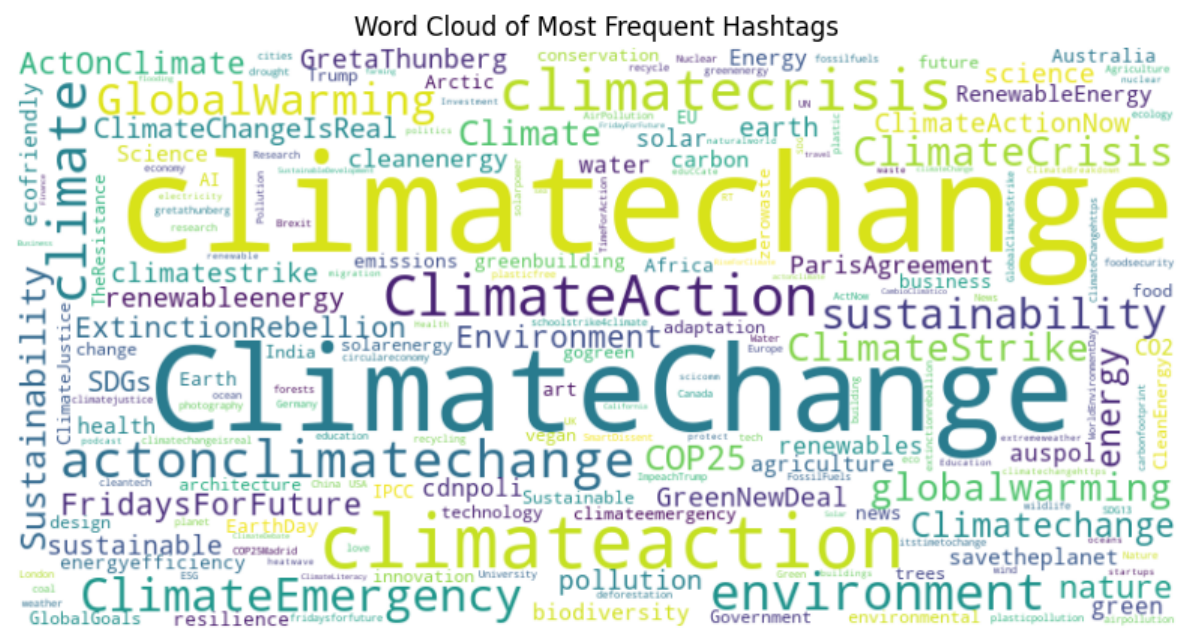}
    \caption{[popular subset] Hashtags from 2019 are mainly associated with positive or neutral discourse on climate change.}
    \label{fig:wordcloud}
\end{figure}

The most prominent hashtags include \#ClimateChange, \#ClimateAction, \#ClimateEmergency, \#GlobalWarming, \#ClimateCrisis, \#ClimateStrike, \#Sustainability, and \#Environment as shown in \Cref{fig:wordcloud}.
This indicates that 2019's twitter discourse was shaped by climate activism.
None of the frequent climate deniers' hashtags show up, i.e. \#climatescam, \#climatehoax and \#climatecult \cite{zak2023colors}.
It appears that in the year 2019, the deniers of climate change, did not use the main-stream hashtags (climatechange) on X discourse.

\begin{figure}[ht]
    \centering
    \includegraphics[width=0.8\linewidth]{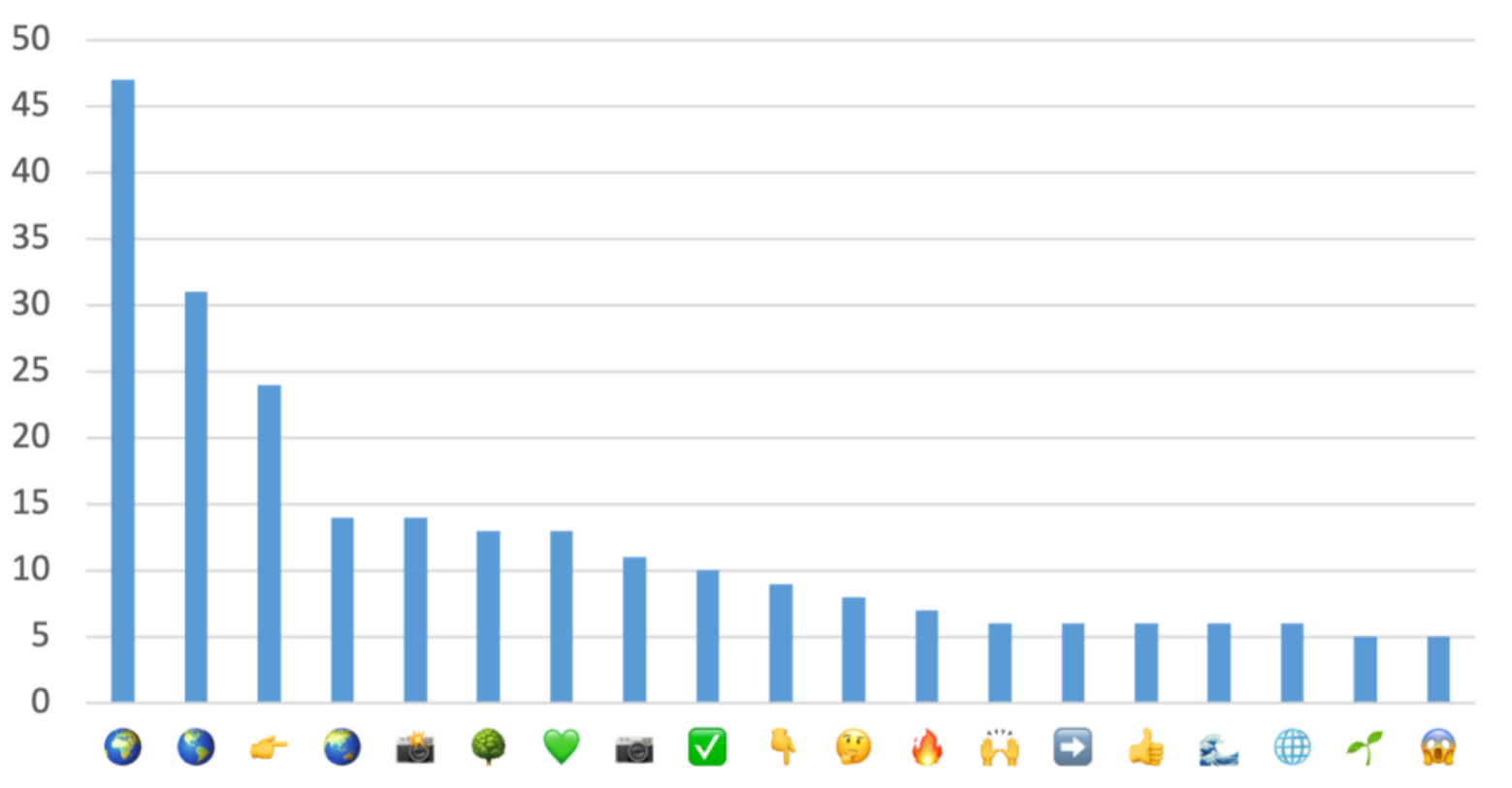}
    \caption{[popular dataset] Emojis in 2019 can mainly be associated with positive or neutral sentiment; only 2 / 19 can be used in a negative sentiment.}
    \label{fig:emoji}
\end{figure}

Similar trends become apparent when analysing the emojis tweeted (compare \Cref{fig:emoji}).
The most popular emojis include representations of the Earth, from the African/European perspective. 
Other perspectives of the earth are also frequent, with the American version coming in second and the Asian/Australian version coming in fourth.
Other frequently used emojis, such as the pointing finger, the camera, and the tree, reinforce messages of urgency, visual documentation, and environmental focus. 
Within the 500 most popular tweets, a tweet contains on average 0.085 emojis. 

\begin{figure}
    \centering
    \includegraphics[width=0.8\linewidth]{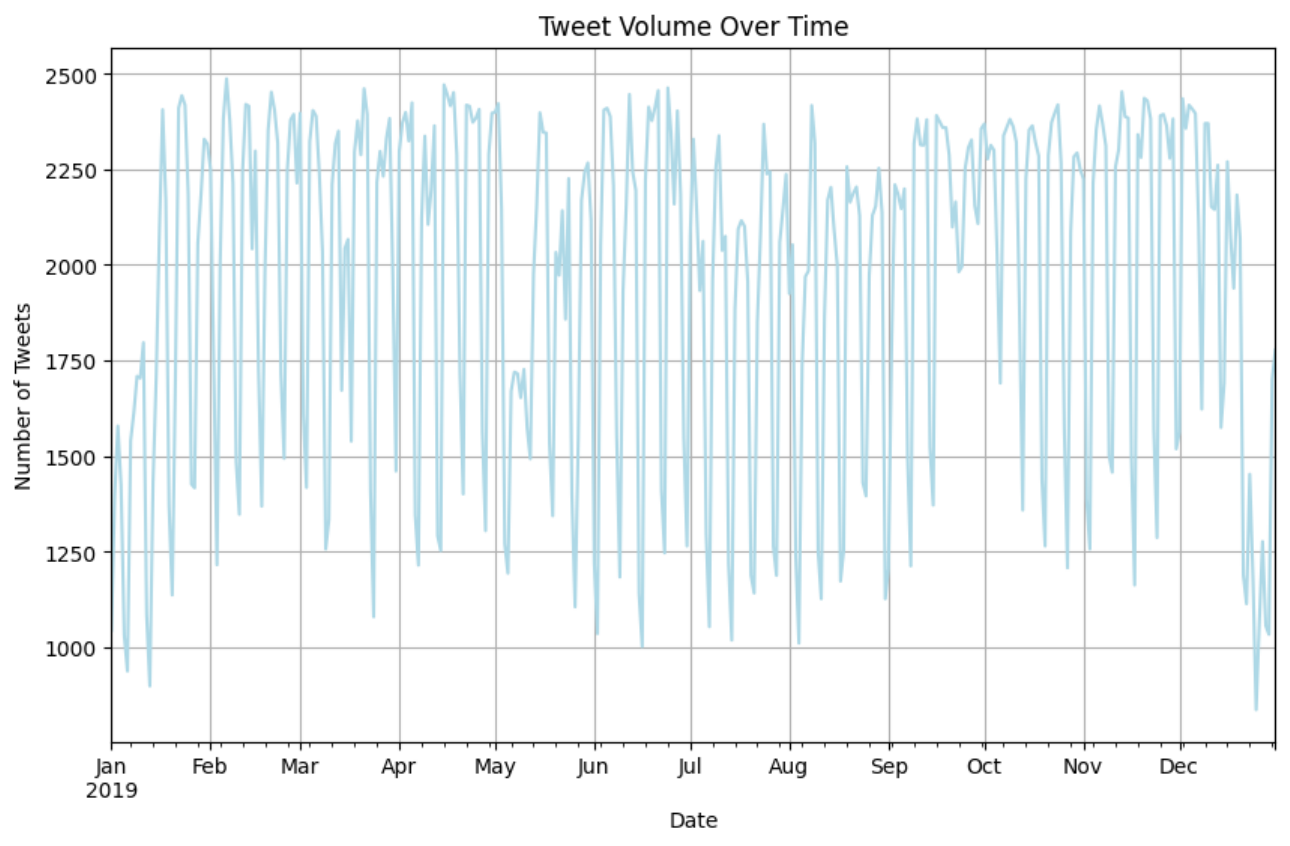}
    \caption{[entire dataset] Tweet frequency varies over the course of 2019 with September being the most consistently busy month.}
    \label{fig:tweet_freq}
\end{figure}

\Cref{fig:tweet_freq} illustrates the tweet volume throughout 2019, displaying daily variations with a recurring pattern of highs and lows across the year. 
A noticeable peak occurs in early February, potentially linked to significant weather events such as a record heatwave in Australia or an extreme cold spell followed by above average temperatures in the US. 
Moreover, this is the month in which the global school strike took place and Fridays for Future gained a lot of attention (compare \cite{mooseder2023social}.
The overall tweet volume remains fairly steady month to month with the exception of September, which might be linked to the UN Climate Action Summit 2019 happening (compare \cite{mooseder2023social}).
The first half of May and the second half of December are the least active times with respect to the climate change discourse.

\begin{figure}[t]
    \centering
    \includegraphics[width=0.8\linewidth]{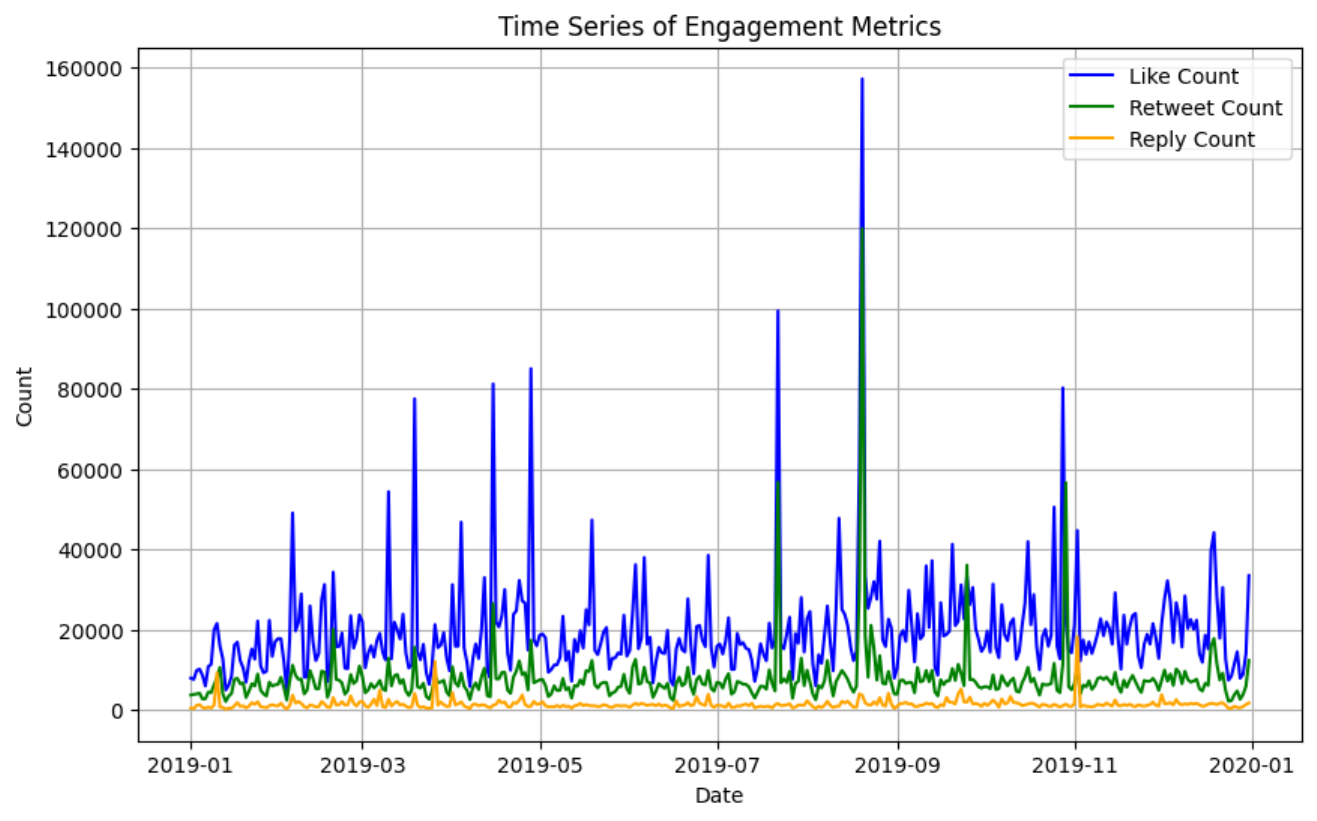}
    \caption{[entire dataset] Engagement varies over the course of 2019 with the largest spike in August.}
    \label{fig:engagement}
\end{figure}

In 2019, the engagement with posted tweets also varies over time, as \Cref{fig:engagement} shows.
Of all metrics, like count consistently leads in volume, with several notable spikes across the year. 
Retweet count exhibits a similar pattern, showing a significant rise alongside like count in August 2019. 
The reply count appears to lag behind the other two metrics, which can intuitively be explained by individuals first needing to see a tweet before they can react to it.

\subsection{Sentiment Analysis}
\begin{figure}[ht]
    \centering
    \includegraphics[width=0.45\linewidth]{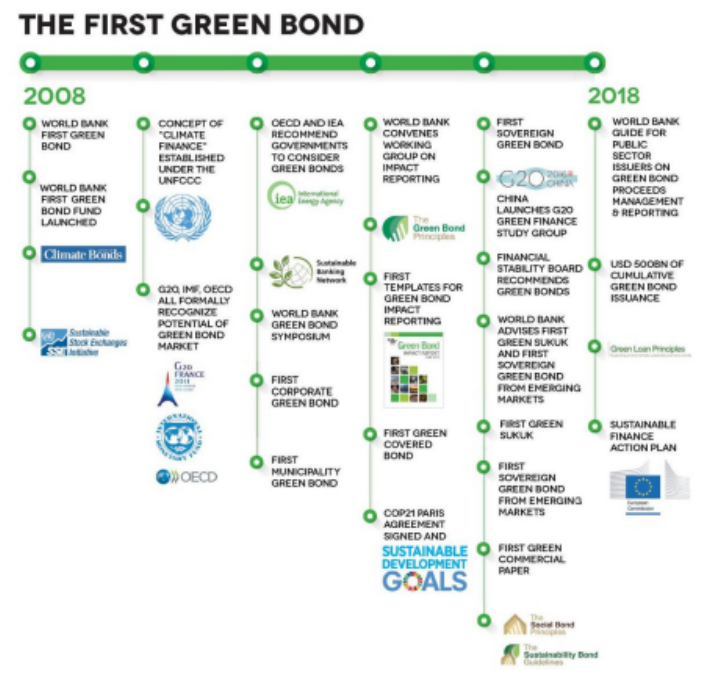}
    \includegraphics[width=0.45\linewidth]{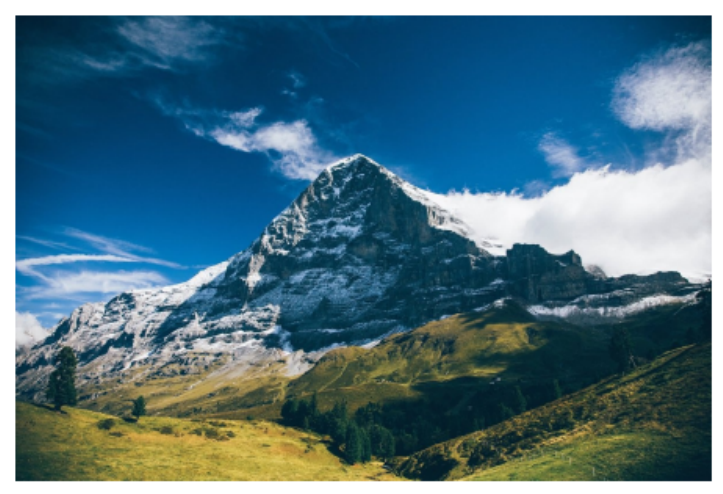}
    \caption{Model errors: Predicting neutral images to contain positive sentiment [Moondream on sentiment subset].}
    \label{fig:pos_neu_pred}
\end{figure}

Model predictions were generally over-positive, as all models' sentiment prediction distributions are more shifted towards the positive label compared to the ground truth (compare \Cref{tab:sent_pred}).
Moreover, the divergence between image and text sentiment becomes already apparent.
Text sentiment prediction had an accuarcy of 67\%, while image sentiment performance was slightly worse with 56\% (GEMINI), 42\% (Lens), and 60\% (Moondream).
\begin{table}[ht]
    \centering
    \begin{tabular}{lccc}
    \toprule    
         & positive & neutral & negative  \\
    \midrule
    \textit{Text GT} & \textit{130} & \textit{198} & \textit{172} \\
    RoBERTa \cite{tan2022roberta} & 143 & 213 & 141 \\
    \midrule
    \textit{Image GT} & \textit{122} & \textit{229} & \textit{149} \\
    Gemini \cite{gemini}* & 212&115 & 151  \\ 
    Lens \cite{lens} & 293& 19&188 \\
    Moondream \cite{moondream} & 183 & 172 & 142 \\ 
    \bottomrule
    \end{tabular}
    \vspace{0.1cm}
    \caption{Sentiment diversity in model predictions, divided by modality [sentiment subset]. GT stands for ground truth. *22 predictions infeasible using GEMINI.}
    \label{tab:sent_pred}
\end{table}
Over all experiments, the misclassified examples were mostly one step away from the true label, e.g "neutral" instead of "positive". 
A reason for the misclassification seems to be that the model does not always recognize sarcasm, e.g. the following tweet text:
\begin{quote}
    Socialists want to do Bad People things to you, like
give you a free education and pay for your healthcare
and have you not die from climate change, by using
Bad People policies a la such wastelands [as] Europe
and Canada.
\end{quote}

is negative according to RoBERTa while just pointing out good actions the socialist party wants.
Thirty images were misclassified as positive instead of neutral by all vision models, with examples shown in \Cref{fig:pos_neu_pred}. 
This shows that neutral content is particularly challenging. 

\begin{figure}
    \centering
        \includegraphics[width=0.9\linewidth]{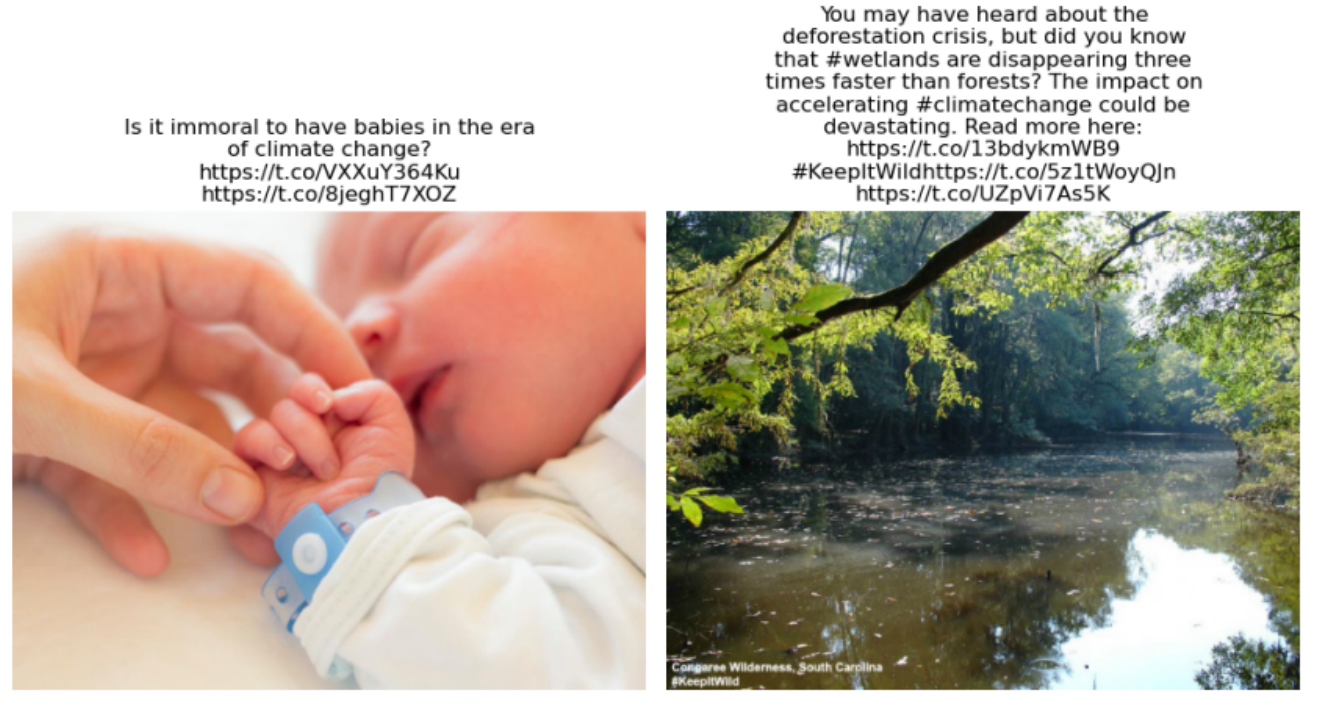} \\
        \includegraphics[width=0.45\linewidth]{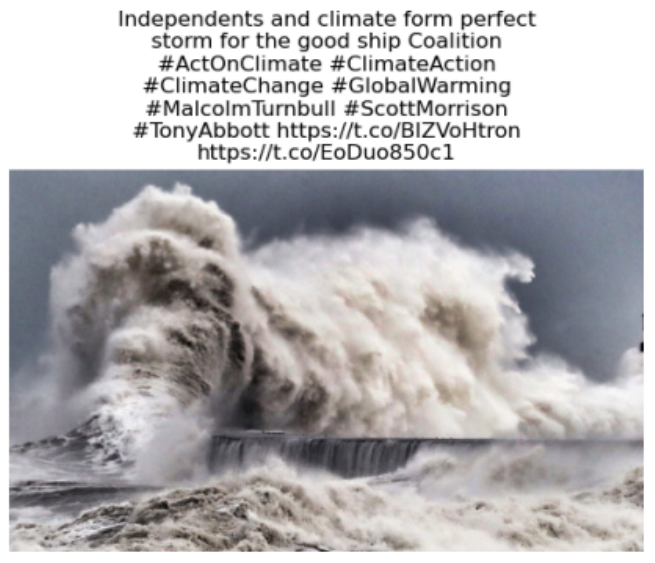}
        \includegraphics[width=0.45\linewidth]{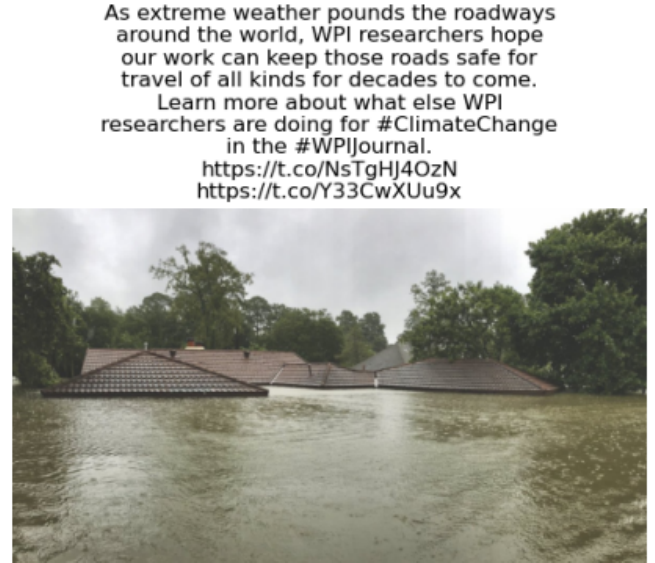}

    \caption{Sentiment divergence with (top row) negative text and positive image, and (bottom row) positive text and negative image [sentiment subset].}
    \label{fig:mismatch}
\end{figure}

\textbf{Sentiment Divergence} is an expected phenomenon on social media. 
The combination of contrasting positive images with negative text is frequently used to provoke thought or to highlight contradictions between appearance and reality, as shown in \Cref{fig:mismatch}.
In 31 instances a positive image sentiment was paired with a negative caption.
Notably, many tweets involved prominent figures and politicians such as Prince William and Kate Middleton or Greta Thunberg, in which the accompanying text often criticizes the individuals, accusing them of hypocrisy among other things. 
Similarly, nature is often depicted in an idyllic state, while accompanying captions frequently criticize human actions that have led to its degradation.
For example, tweets questioning ethical or life decisions might pair an innocent or joyful image with a critical or pessimistic caption, such as a discussion on the morality of bringing a child into a world faced with significant challenges.
The bottom row of \Cref{fig:mismatch} shows tweets where a negative image sentiment is paired with a positive text sentiment. 
In these examples, images depicting environmental degradation, such as melting icebergs or deforestation, are accompanied by captions that express hope, resilience, or a call to action. 
Additionally, another prominent category in this subset involves images of demonstrations or protests, where the image sentiment is often negative, reflecting the tension inherent in such events. 
However, the accompanying captions frequently convey messages of optimism and solidarity which offer hope for change despite the challenging circumstances depicted.


\section{Interactive GUI}
To better interpret visual narratives, we have created a Graphical User Interface (GUI), which provides a large preview of the tweet's image along with the most important tweet information and the corresponding results.
Given that we want to make the code open source, we selected a framework which is free and provides stable long-term versions. 
We also preferred a widely adopted framework which exists alongside plentiful resources and tutorials, as this allows further development by the community. 
Finally, we opted for the Vue \cite{vue} framework due to its intuitive syntax, simplicity, and favorable learning curve. 
For our front-end and design, we used plain CSS \cite{css} added by some Bootstrap \cite{bootstrap}.
Further details can be found in  \Cref{sec:gui}.

\section{Discussion and Conclusion}
We analysed 700k climate change-related tweets using image classification, object detection, and sentiment analysis. 
Frequency analysis indicates that X's discourse aligns with related events and can thus provide insights into their perception.
Automated sentiment analysis achieved a maximum accuracy of 67\%, underscoring a need for further model improvement and sentiment dataset.
The detailed differentiation between emotions would allow for additional insights.
The tweets with divergent image and text sentiment are especially interesting in our opinion, since in this case, the use of multi-modal models is less effective than in tweets where both modalities belong to the same sentiment.
This highlights the need for an interactive GUI to facilitate deeper evaluation of model predictions.
By making our code and tools open-source, we aim to lay the groundwork for advancing AI-driven climate communication research.
\bibliography{main}
\bibliographystyle{plain}

\newpage

\appendix
\section{Gemini prompt}
For this model we used a return structure which allowed us to further interpret the sentiment assessment by analysing the confidence score and the model’s explanation. 
However, this prompting technique did not work for all tested models, as the majority merely returned the sentiment regardless of the provided desired JSON return structure.

\begin{quote}
What is the sentiment of the image?
Please classify the image’s sentiment
into positive, negative, and neutral
and provide a confidence score in
[0,1] as well as a short explanation.
Structure your answer in the same json
format as this example and do not add
any additional information:
{"sentiment": "XX", "confidence": XX,
"explanation": "XX"}
\end{quote}
GEMINI prompt.

\section{Image Classification and Object Detection}
\label{sec:object}
\begin{figure}[ht]
    \centering
    \includegraphics[width=0.95\linewidth]{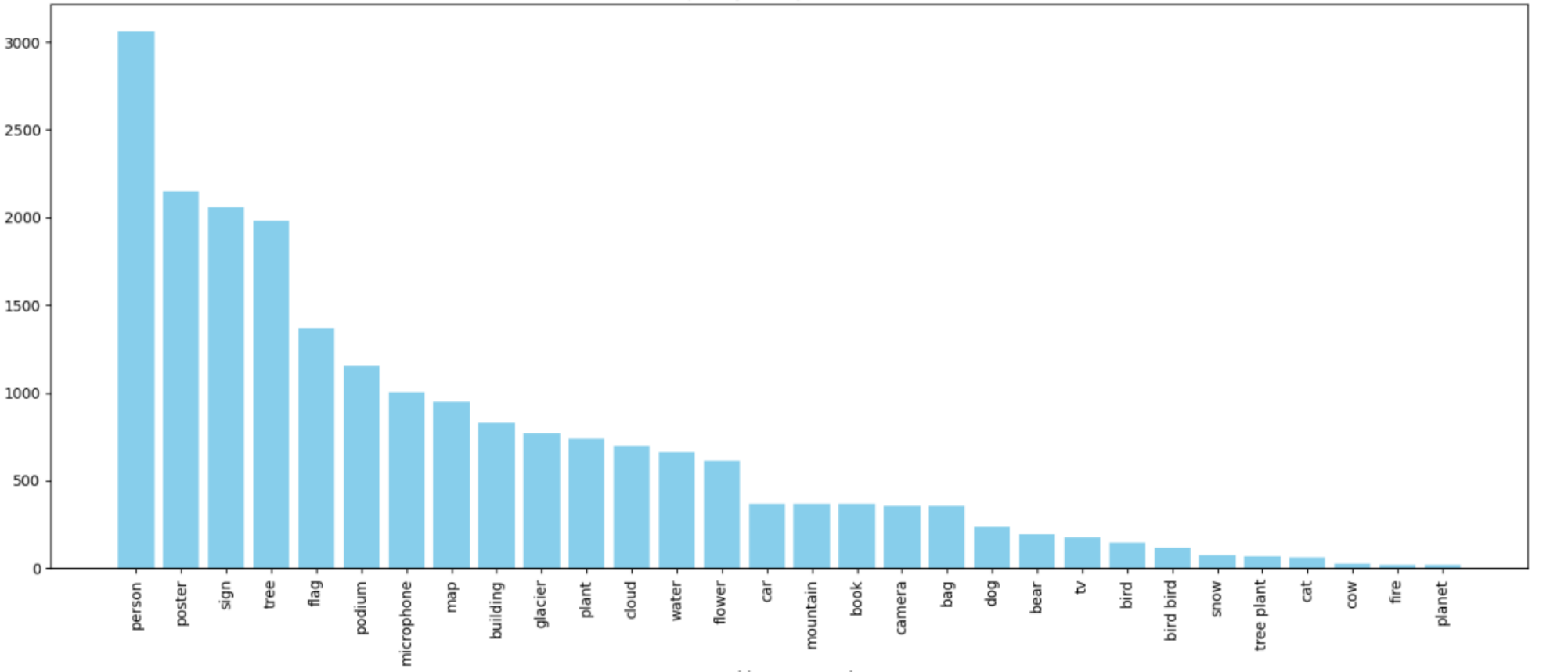}
    \caption{Object detection using GroudingDINO \cite{liu2025grounding} [popular subset].}
    \label{fig:groundingdino}
\end{figure}

We focus our report on object detection, as this is also the task less discussed on other works.
In \Cref{fig:groundingdino} we highlight the most frequently detected objects.
While news-worthy terms are included (e.g. glacier, fire, planet) and common object in public discourse (e.g. podium, microphone, flag), the data reveals also common objects (e.g. book, tv, cat, dog).
This indicates that individuals share not only public content but also their own climate change experiences, even in the most popular subset.

\section{GUI}
\label{sec:gui}

We host both the front-end and back-end on a dedicated Linux server.
In order to manage large amounts of data within reasonable processing times, we implemented a Python Flask \cite{flask}. We chose Flask since it allows users to easily
upload output files as Excel documents, which can then be processed with Python’s Pandas library \cite{pandas} and it thus provides great flexibility for data aggregation, reformatting, and custom function applications.

\subsection{Filters and HTTP-Requests}
To be able to apply text-based filters to the tweets, we came up with a modular filter system, where each column has a checkbox and an adjacent input field. 
Upon selecting the desired column(s) and filling out their text-based inputs, they get concatenated with the column-name to an HTTP-Request, that the back-end is serving. 
Next, the back-end responds with a list of only the entries that the filter criteria apply to.
To avoid downloading too much data when having many
search results, a second HTTP-request is then triggered to just
get the information of the currently shown picture and show it
in the GUI. 
The same idea applies to the images themselves,
out of the entry list corresponding to the search criteria, only
the current nine previews get downloaded and displayed. 

\subsection{Design}
Regarding the appearance, we aimed to develop a neutral, clean, and intuitive GUI focused on analyzing our findings. 
To achieve this, we adopted a simple and structured design that presents our results objectively. 
The main analysis page is organized vertically, featuring images, navigation elements, results, and filtering options in that order.
Additionally, we prioritized a pleasant user experience by creating supplementary pages such as the \textit{Home} and \textit{About} pages. 
These pages contribute to a professional website feel, enhancing users’ perception of our tool and incentivizing them to stay on our page for longer. 
Furthermore, the FAQ section provides guidance for users and can be easily updated with future additions and thus adapted to user feedback.

\end{document}